%% file: root.tex
\documentclass[letterpaper, 10 pt, conference]{ieeeconf}  

\IEEEoverridecommandlockouts                              

\usepackage{graphics} 
\usepackage{epsfig} 
\usepackage{mathptmx} 
\usepackage{times} 
\usepackage{amsmath} 
\usepackage{amssymb}  
\usepackage{cite}
\usepackage{svg}
\usepackage{amsfonts}
\usepackage{algorithmic}
\usepackage{textcomp}
\usepackage{xcolor}
\usepackage{multirow}
\usepackage{booktabs}

\title{\LARGE \bf
Towards Latency-aware 3D Streaming Perception\\ for Autonomous Driving}
\author{Jiaqi Peng$^{1,2}$, Tai Wang$^{2}$, Jiangmiao Pang$^{2}$ and Yuan Shen$^{1,2}$
\thanks{$^{1}$Jiaqi Peng, Yuan Shen are with the Department of Electronic Engineering, Tsinghua University.
        {\tt\small shenyuan\_ee@tsinghua.edu.cn}
        }%
\thanks{$^{2}$Jiaqi Peng, Tai Wang, Jiangmiao Pang and Yuan Shen are all with the Shanghai AI Laboratory.
        }%
}

\begin{document}

\maketitle
\thispagestyle{empty}
\pagestyle{empty}

\input{0.abstract}
\input{1.intro}
\input{2.related}
\input{3.study}

\input{4.method}
\input{5.experiment}
\input{6.conclusion}



\bibliographystyle{IEEEtran}
\bibliography{root}

\end{document}

%% file: 0.abstract.tex
\begin{abstract}
Although existing 3D perception algorithms have demonstrated significant improvements in performance, their deployment on edge devices continues to encounter critical challenges due to substantial runtime latency. 
We propose a new benchmark tailored for online evaluation by considering runtime latency.
Based on the benchmark, we build a Latency-Aware 3D Streaming Perception (LASP) framework that addresses the latency issue through two primary components: 
1) latency-aware history integration, which extends query propagation into a continuous process, ensuring the integration of historical feature regardless of varying latency; 
2) latency-aware predictive detection, a module that compensates the detection results with the predicted trajectory and the posterior accessed latency.
By incorporating the latency-aware mechanism, our method shows generalization across various latency levels, achieving an online performance that closely aligns with 80\% of its offline evaluation on the Jetson AGX Orin without any acceleration techniques.

\end{abstract}

%% file: 1.intro.tex
\section{INTRODUCTION}

3D perception is an essential capability for autonomous vehicles and provides the foundation for subsequent prediction and planning~\cite{ma2022vision,mao20233d}. The past few years have witnessed the rapid advancement of 3D perception algorithms~\cite{wang2021fcos3d,huang2021bevdet,li2022bevformer}.
In particular, one of the most popular settings, \emph{i.e.}\cite{park2022time,wang2023exploring}, only using multiple cameras, can achieve performance that is comparable with LiDAR-based methods~\cite{yin2021center,lang2019pointpillars} with effective BEV-based paradigms~\cite{li2022bevformer,park2022time}.
However, behind this progress, it is noteworthy that the improvement of 3D perception accuracy typically sacrifices the inference efficiency, leading to lagged predictions in practice. Deploying these algorithms on edge devices still faces critical challenges.

\begin{figure}[t]
\begin{center}
  \includegraphics[width=\linewidth]{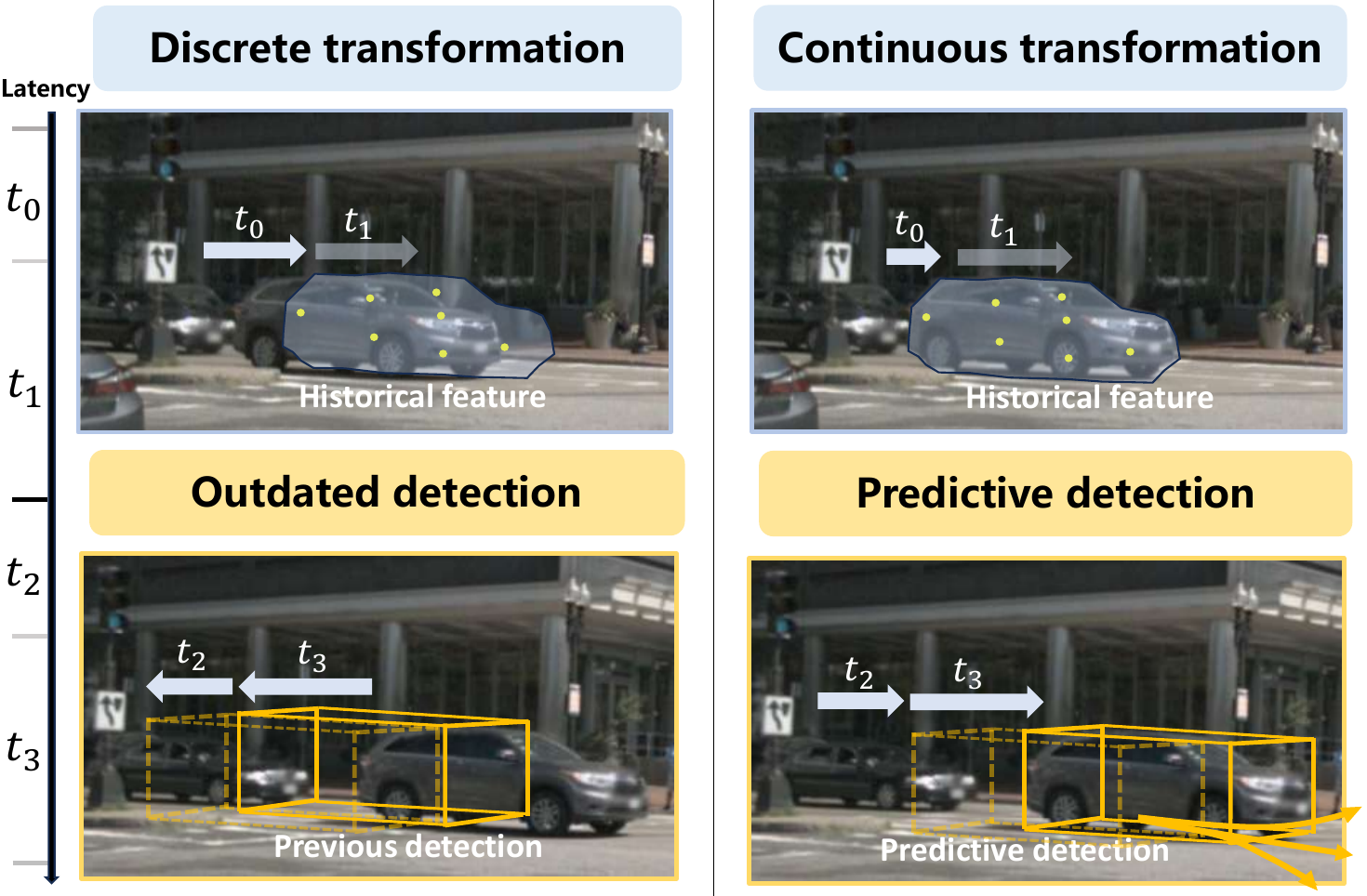}
  \vspace{-0.8cm}
    \end{center}
    \caption{
    When deployed on edge devices, random latency introduces two challenges: 1) irregular historical frames, which compromise the effectiveness of discrete transformation; 2) time-lagged prediction, leading to mismatches between the predicted states and real-time detection outcomes.
    }
\label{fig:teasor}
\vspace{-0.5cm}
\end{figure}

Given the progress of lightweight framework design and optimization~\cite{lee2019energy,sun2024multi}, we target another aspect of this problem, streaming perception, a setting first introduced in \cite{li2020towards}. Rather than prioritizing the lightweight design and maximized processing speed, streaming perception requires algorithms to predict results that account for the inevitable latency inherent in real-time processing.
While this problem has been extensively studied in the 2D domain~\cite{yang2022real,li2023longshortnet,he2023damo,ghosh2021adaptive}, it was hardly studied in 3D perception.
The only work~\cite{wang2023we} builds a benchmark for camera-based 3D streaming detection and proposes baseline approaches similar to those in 2D\cite{yang2022real}.
However, we find this work still falls short of addressing the complexities of real-world applications: 1) Most state-of-the-art algorithms rely on historical frames as input, whereas the benchmark only considers the single-frame case, 2) The practical case has random latency, leading to irregular time steps of historical frames and lagged prediction for unpredictable timing.

To bridge this gap, we first establish a benchmark tailored for online evaluation by considering runtime latency.
This benchmark allows for the customization of input data streams and supports online evaluation under various latency conditions.
Our investigation reveals the limitations of current 3D perception algorithms in handling irregular historical frames and providing online output.

Based on this benchmark, we propose the Latency-aware 3D Streaming Perception (LASP) framework, involving latency-aware designs for both history integration and predictive detection. 
Specifically, in contrast to previous works assuming a linear motion-based mapping across the discrete interval, we develop a continuous formulation and derive a linear ordinary differential equation to model the temporal evolution of object query over continuous time, enabling the model to accommodate irregularly spaced historical frames.
Furthermore, to enable latency-aware prediction, we decompose the problem into a latency-agnostic trajectory prediction, which is performed by a lightweight network running concurrently with the detection process, and a latency-compensated detection, which adjusts the detection results based on the predicted trajectory and the posterior latency.

Our method achieves online performance that closely aligns with 80\% of its offline evaluation on the Jetson AGX Orin without utilizing any acceleration techniques and outperforms other algorithms deployed on more powerful platforms. 
Furthermore, it surpasses the performance of models accelerated by TensorRT, demonstrating that our approach provides superior results even without specialized hardware optimizations, highlighting the advantages of streaming perception research.

%% file: 2.related.tex
\section{Related Work}

\subsection{Camera-based 3D Perception}
The field of camera-based 3D object detection has witnessed rapid advancements towards efficiency and accuracy. Based on feature representation, existing methods can be broadly classified into BEV-based methods and query-based methods. \textbf{BEV-based methods}\cite{park2022time,huang2021bevdet,li2023bevdepth,liu2023sparsebev,li2022bevformer} aim to construct a unified BEV space by leveraging estimated depths or attention layers to project multi-view images into the bird’s-eye view, and then employ a BEV-based detector on top of the unified BEV map to detect 3D objects. \textbf{Query-based paradigms}\cite{wang2023exploring,liu2024ray} resort to 3D object queries that are generated from the bird’s-eye view and Transformers where cross-view attention is applied to object queries and multi-view image features. They learn global object queries from the representative data, then feed them into the decoder to predict 3D bounding boxes. Current state of art methods, such as\cite{huang2022bevdet4d,lin2022sparse4d,liu2023petrv2}, only using multiple cameras, can achieve performance that is comparable with LiDAR-based methods. However, when integrating historical features, these methods typically employ discrete transformations based on ego pose and velocity, often neglecting timestamp information in modeling object temporal motion\cite{qing2023dort}.

\subsection{Streaming Perception}
The concept of streaming perception originated from 2D image detectors in autonomous driving\cite{li2020towards}, where algorithms are evaluated based on both their accuracy-latency trade-off\cite{li2021predictive,li2023pvt++}. Given results discrepancies caused by latency, Streamer\cite{li2020towards} proposed a meta-detector to mitigate this issue by employing a Kalman filter, dynamic scheduling, and asynchronous tracking. After discovering that real-time detectors can narrow the performance gap between streaming and offline perception, StreamYOLO\cite{yang2022real} restructured the task to predict the results of the next frame based on the current state and introduced a Dual-Flow Perception module to fuse features from the previous and current frames. Subsequent algorithms\cite{li2023longshortnet,he2023damo} have also focused on incorporating historical features into real-time detectors. 
In the absence of ego pose information, 2D streaming perception must model not only the motion of objects but also the movement of the camera\cite{aharon2022bot}. 
In 3D streaming perception tasks, while global physical attributes can be aligned using ego-pose data, modeling object motion remains difficult. Moreover, in real-world scenarios, especially where the computational power of edge devices is limited, effectively integrating off-the-shelf historical features to predict future states becomes crucial\cite{gu2023vip3d,zhou2023query}.

\begin{figure*}[t]
\begin{center}
  \includegraphics[width=\textwidth]{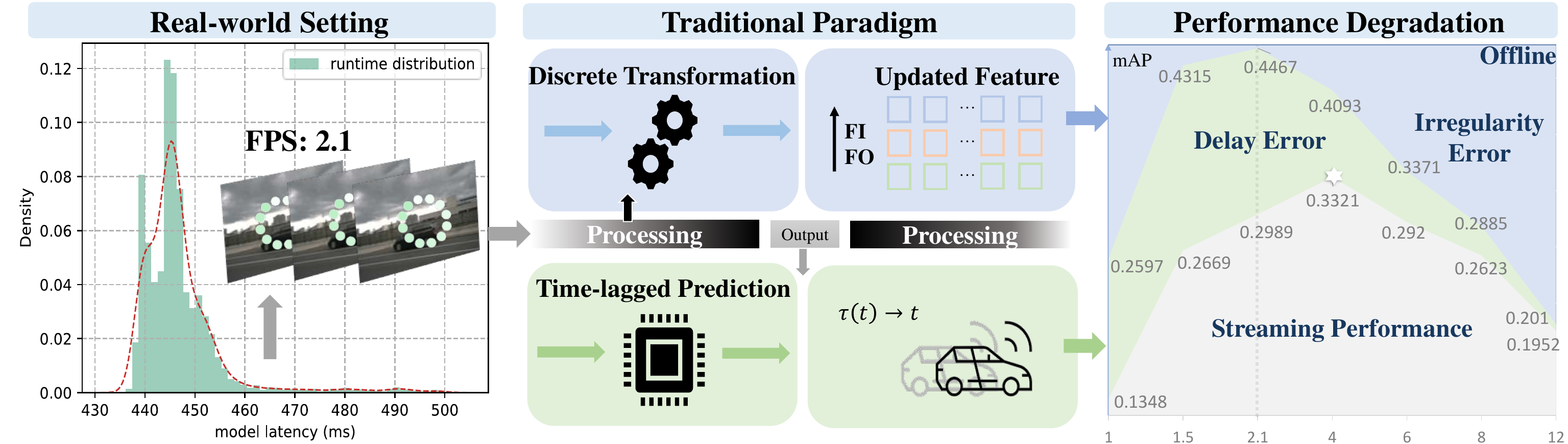}
  \vspace{-0.9cm}
\end{center}
  \caption{
  The left figure illustrates the runtime distribution on the edge device. We sampled runtime of each frame, applying discrete transformations to process historical feature and directly using time-lagged predictions with delay. The right figure demonstrates the performance degradation compared to offline results, highlighting the limitations of current algorithms in handling historical frames at various intervals and providing online output.
}
\label{fig:pipeline}
\vspace{-0.5cm}
\end{figure*}

%% file: 3.study.tex
\section{Streaming Benchmark and Analysis}
In this section, we introduce a streaming benchmark designed to enhance the real-time perception. Then we analyze the challenges of streaming task when evaluating current camera-based 3D perception algorithms.

\subsection{Streaming Benchmark}
Following the 2D streaming perception paradigm\cite{li2020towards}, we conduct online evaluation by considering runtime latency. 
At any continuous time $t$, a data stream is characterized by a set of sensor observations and timestamps, denoted as 
\begin{align}
D_{\text{s}} = \{(x_i,t_i)\mid t_i\leq t\}
\end{align}
where each tuple $(x_i, t_i)$ encapsulates an observation $x_i$ along with timestamp $t_i$, which allows us to evaluate the algorithm's ability to integrate historical information. The intervals between data streams are influenced by random latency $\tau$, resulting in arbitrary timestamps for each frame.

In a real-time, streaming context, the streaming algorithm $f$ needs to generate the state predictions after random latency $\tau$. To facilitate the evaluation on the perception of future continuous states, we define a series of query times to examine future states at these discrete points, which can be set from the start of the next frame until the end of its execution, at which evaluations are performed: 
\begin{align}
\{t_j\}_{j=0}^N = \{ t_j  \mid t + \tau\leq t_j < t + 2\tau \}
\end{align}
The algorithm is required to immediately return perception results without relying on any time-consuming operation:
\begin{align}
\{\hat{y}_j\}_{j=0}^{N}=f(\{(x_i,t_i)\mid t_i\leq t\},\{t_j\}_{j=0}^N)
\end{align}
We can assess these time points using metrics $L$ commonly used in 3D perception benchmarks, based on the ground truth for each frame:
\begin{align}
L_{\text{s}}=L(\{y_i,\hat{y}_i\}_{j=0}^N)
\end{align}

The camera sweeps from the nuScenes dataset\cite{caesar2020nuscenes} are leveraged as streaming input to evaluate the continuous time prediction ability of each algorithm in a fine granularity. Our benchmark allows for customization of input data streams and intervals based on the latency of hardware platforms.

\subsection{Performance Analysis}
The errors in streaming mainly stem from two main sources: irregular intervals of historical data and time-lagged prediction caused by latency. Fig. \ref{fig:pipeline} illustrates the model's performance degrades under different latency intervals. 

Since the original model was trained using a fixed frame rate, the first performance drop is observed when the input deviates from this rate. Even equipped with explicit physical alignment modules or temporal embeddings, it still lacks generalization in aligning historical information across varying frame rates. Therefore, precise alignment becomes difficult when time intervals between frames are irregular.

The second performance drop is due to increasing latency, where larger latency results in a more pronounced delay in predictions, leading to a significant performance gap compared to offline results. Notably, \textbf{the temporal delay in the results can be up to twice the model latency}: by the time the model processes the latest frame, the data corresponds to the previous delay interval, and this result may be ascribed to the next delay interval's timestamp. We observed a significant degradation in metrics when operating at perception frequency slower than 2Hz. With just a 0.5-second time interval between frames, when evaluated from key frames to all sweeps, the algorithm experienced a 14.8\% drop in mAP.

%% file: 4.method.tex
\section{Method}

\subsection{Overall Architecture}
As illustrated in Fig. 3, our method builds on end-to-end sparse query-based 3D object detectors~\cite{wang2023exploring}, incorporating a continuous temporal alignment mechanism to integrate irregular historical feature and a predictive detection head for forecasting future object states.

We employ a conventional 2D backbone\cite{he2016deep,lee2019energy} as the image encoder to extract semantic features from multi-view images. A memory queue~\cite{wang2023exploring} stores irregular streaming information, such as timestamps and context embeddings. Subsequently, the historical object queries are continuously aligned to the current timestamp, together with the features extracted from the latest frame are fed into the transformer decoder to facilitate spatial-temporal interaction. Finally, an intention-guided head forecasts the object state and its evolution over a defined period. 

\begin{figure*}[t]
  \centering
  \includegraphics[width=\textwidth]{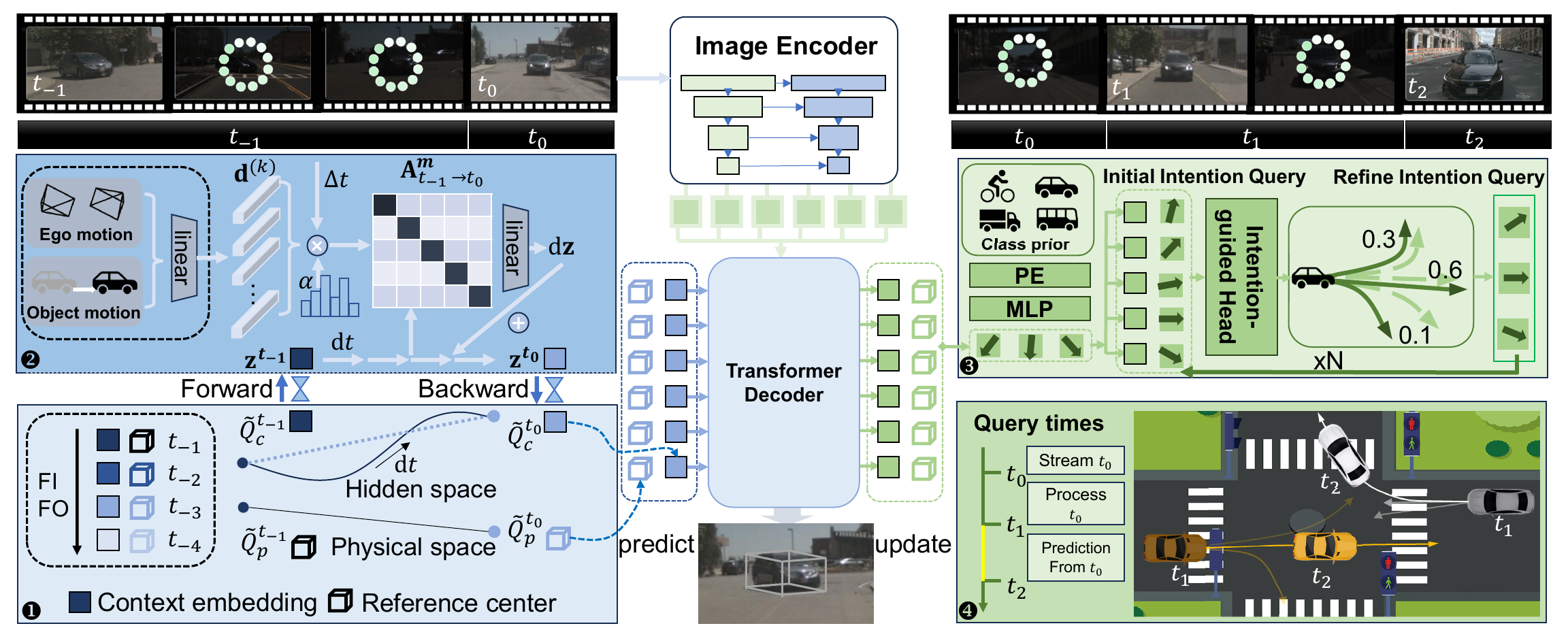}
  \vspace{-0.7cm}
  \caption{Our method builds on end-to-end sparse query-based 3D object detectors: 
  1) We maintain a memory bank containing historical query context embedding and reference centers, which are aligned to current time $t_0$ and fed into the transformer decoder.
  2) Assuming that the hidden state follows a motion-aware linear transition over a small time step $\text{d}t$, we integrate it over time to obtain the hidden state at $t_0$.
  3) The updated queries are merged with class prior and passed through a intention-guided head to forecast the future trajectories which are re-encoded for further refinement.
  4) Data arriving at $t_0$ is processed until $t_1$, generating trajectories that are used to compensate for the movement from $t_1$ to $t_2$.
  }
  \label{fig:qualitative}
  \vspace{-0.5cm}
\end{figure*}

\subsection{Latency-aware History Integration}
Latency-aware history integration is designed to integrate irregular interval streaming data. It can implicitly update the object state to any continuous timestamp according to the frame interval and motion information recorded in the memory queue. 

To model object movement, object queries are employed, where each query consists of two components: the query reference center and the query context embedding, denoted as: $\tilde{Q}^t=(\tilde{Q}^t_p, \tilde{Q}^t_c)$.
When moving to the next timestamp, the object may experience a state change.
These changes can easily be expressed in physical space and mapped through an explicit coordinate transformation.
Assuming we have a list of observation times $\mathcal{T}=\{t_0,t_1,...,t_N\}$ that occur at irregular intervals, we aim to estimate the state of the object at time $t$. The historical observation time is 
\begin{align}
\tau(t)\in \{t'\in\mathcal{T}, s.t. t' < t\}
\end{align}
By applying ego pose transformation $E_{\tau(t)}^t$ from $\tau(t)$ to $t$ and the object’s own transition, the query 3D center at time $t$ is
\begin{align}
\tilde{Q}_p^t=E_{\tau(t)}^t\cdot Q_p^{\tau(t)}+v^{\tau(t)} \Delta t
\end{align}
However, the query context embedding cannot be explicitly transformed by these motion attributes. As a result, some methods\cite{liu2024ray, wang2023exploring} adopt motion-aware linear layers $\mathbf{A}_{\tau(t) \rightarrow t}^\mathbf{m}$ to transform the query embedding:
\begin{align}
    \tilde{Q}^t_c = \mathbf{A}_{\tau(t) \rightarrow t}^\mathbf{m} \tilde{Q}^{\tau(t)} \in \mathbb{R}^{M}
\end{align}

Unlike discrete hidden state formulations above like RNNs\cite{elman1990finding,choi2018mime}, to extend the query propagation to continuous time transformation, we first use a non-linear layer to convert the query embedding $\tilde{Q}^t_c$ into a hidden space: $\mathbf{z}^t=\phi(\tilde{Q}^t_c)\in\mathbb{R}^{M}$. And then we assume that in small time step $\text{d} t$, the hidden state $\mathbf{z}^t$ also follows motion-aware linear transition, which is governed by a linear ODE\cite{schirmer2022modeling, revach2022kalmannet}:
\begin{align}
\text{d}\mathbf{z}=\mathbf{A}_{\tau(t) \rightarrow t}^\mathbf{m} \mathbf{z}\text{d}t
\end{align}
with time-invariant transition matrix $\mathbf{A}_{\tau(t) \rightarrow t}^\mathbf{m} \in \mathbb{R}^{M\times M}$. The motion attributes $\mathbf{m}=(E_{\tau(t)}^t,v,\Delta t)$ constitute all the information in the physical world coordinate transformation. Consequently, the transformation matrix $\mathbf{A}_{\tau(t) \rightarrow t}^\mathbf{m}$ in the hidden space should inherently be associated with these motion attributes, reflecting the dynamic nature of the object. The solution to the ODE at time $t$ is
\begin{align}
\mathbf{z}^t=\textbf{exp}(\mathbf{A}_{\tau(t) \rightarrow t}^\mathbf{m}(t-\tau(t)))\mathbf{z}^{\tau(t)}
\end{align}
where \textbf{exp} denotes the matrix exponential.
Similar to approaches that have been used in dynamic state-space models\cite{karl2016deep, fraccaro2017disentangled}, we employ the locally linear transition model to increase modeling flexibility. The transition matrix $\mathbf{A}_{\tau(t) \rightarrow t}^\mathbf{m}$ can be expressed as a weighted sum of $K$ parameterized basis matrices:
\vspace{-0.2cm}
\begin{align}
\mathbf{A}_{\tau(t) \rightarrow t}^\mathbf{m}=\sum_{k=1}^K \alpha^{(k)}\mathbf{A}^{(k)},  \text{  with  }\mathbf{\alpha}=\omega_\alpha(\mathbf{m})
\end{align}

While the matrix exponential operation can be computationally expensive, we mitigate this by directly parameterizing the basis matrices $\mathbf{A}^{(k)}$ using their eigenvalues and eigenvectors, which enables a change of basis.
Specifically, we assume basis matrices $\mathbf{A}^{(k)}$ that share the same orthogonal eigenvectors. For all $k \in\{1,...,K\}$, we have $\mathbf{A}^{(k)}=\mathbf{E}\mathbf{D}^{(k)}\mathbf{E}^T$, where $\mathbf{D}^{(k)}$ is a diagonal matrix whose $i$-th diagonal entry is the eigenvalue of $\mathbf{A}^{(k)}$ corresponding to the eigenvector in the $i$-th column of $\mathbf{E}$. In order to correlate the transformation matrix with motion attributes $\mathbf{m}$, the diagonal elements in $\mathbf{D}^{(k)}$ are predicted from the motion attributes $\mathbf{m}$: $\mathbf{D}^{(k)} = \text{diag}(\mathbf{d}^{(k)})\text{, }\mathbf{d}^{(k)} = \omega_d(\mathbf{m})$.
Thus, the matrix exponential simplifies to the element-wise exponential function:
\begin{align}
\mathbf{z}^t=\mathbf{E} \exp((t-\tau(t))\sum_{k=1}^K\alpha^{(k)}\mathbf{D}^{(k)})\mathbf{E}^T\mathbf{z^{\tau(t)}}
\end{align}

Despite that our method incorporates time integration into the module, it remains computationally efficient when compared to conventional methods. For consistency, alignment is also adopted into current object queries. After a nonlinear transformation, the hidden state $\mathbf{z}^t$ is mapped back to the object query embedding. Subsequently, queries from last frame are concatenated with randomly initialized queries and fed into the multi-layer transformer which is connected to the predictive detection head to predict the state of each object.

\subsection{Latency-aware Predictive Detection}
In real-time streaming scenarios, the model is required to output a series of future state predictions that account for latency without introducing additional computational delays. 
Drawing inspiration from model predictive control\cite{rawlings2017model}, we extend the 3D detection model by incorporating model predictive detection module to enable latency-agnostic trajectory prediction that remain effective throughout its running time.

To accurately identify an agent's motion model, we introduce learnable intention queries designed to reduce uncertainty in future state predictions\cite{shi2024mtr++,shi2022motion}. 
Specifically, for each category, we generate $K$ representative intention points $I_a \in \mathbb{R}^{K \times 2}$ for every agent, by applying the k-means clustering algorithm to the endpoints of ground truth trajectories within the training dataset, and transform these points into the ego coordinate $I_e^0$. Each intention point implicitly represents a specific motion mode, encompassing both direction and speed. For a given agent's intention points, we define each intention query as the learnable positional embedding corresponding to a particular intention point:
\begin{align}
E^0[k] = \text{MLP}(\text{PE}(I_e^0[k])) 
\end{align}
where $ k \in \{1, \ldots, K\}$ and $E \in \mathbb{R}^{K \times M}$. Here, $\text{PE}(\cdot)$ represents the sinusoidal positional encoding. 
We incorporate the information from intention queries based on the class predictions generated by detection queries. 
Additionally, we introduce a lightweight intention-guided head to fuse with updated detection queries $Q_c$ and predict multimodal trajectories $I_e$:
\begin{align}
I_e^{(l)}[k] = \text{Dec}(Q_c^{(l-1)} + E^{(l-1)}[k])
\end{align}
In the subsequent decoding layer $l$, we re-encode the endpoint of the trajectory $I_e^{(l-1)}$ from the previous layer $l-1$ to obtain a more accurate intention query, iteratively refining the output. Supervision is applied to the trajectories at each layer throughout this process.

In an online setting, data is streamed at time \( t_0 \). By time \( t_1 \), the system completes processing and generates predicted future trajectories. After accounting for posterior latency, the model progressively adjusts the detection results by predicted trajectories during the time the system was blocked, from the start of the next frame $t_1$ until the end of its execution $t_2$, ensuring accurate real-time predictions.

%% file: 5.experiment.tex
\section{Experiment}
\subsection{Experiment Setup}
\noindent \textbf{Datasets}: We evaluate our approach on the extended nuScenes dataset\cite{caesar2020nuscenes}, with annotations at a rate of 12Hz, which is faster than the inference speed of the majority of camera-based 3D detectors on edge devices, thus can support investigating the streaming performance on various latency. Each scene is a 20 seconds video with about 240 frames. It offers 700 training scenes, 150 validation scenes, and 150 testing scenes with 6 camera views. It consists of 165280 samples for training and 35364 samples for validation, 6 times the amount of original nuScenes dataset. 
We benchmark the streaming algorithm $f$ by comparing its predictions at time $t_j$ to the ground-truth $y_j$. Then we can utilize metrics commonly used in 3D perception: the nuScenes Detection Score(NDS), mean Average Precision(mAP), and 4 kinds of True Positive(TP) metrics,  including average translation error(ATE), average scale error(ASE), average orientation error(AOE), average attribute error(AAE). 

\noindent \textbf{Implementation Details}: We conduct experiments with ResNet50\cite{he2016deep} backbones with pre-trained weights on ImageNet, and train by AdamW optimizer with a batch size of 8. The base learning rate is set to 4e-4, following a cosine annealing learning rate schedule. Due to the excessive length of the video (up to 240 frames), we divide each video into 6 segments for training, introducing more diverse starting scenes. To simulate the pattern of data stream sampling in real-world, we introduce random perturbations to the frame interval around the latency and train the customized input sequences accordingly. In the historical integration, we have chosen 10 basis matrices for locally linear transition. In the predictive detection module, we have set 6 intention queries for each agent. Experiments are conducted on Nvidia Jetson AGX Orin for edge devices inference and RTX 4090 GPUs for training.

\subsection{Main Results}

\begin{figure}[t]
\begin{center}
  \includegraphics[width=\linewidth]{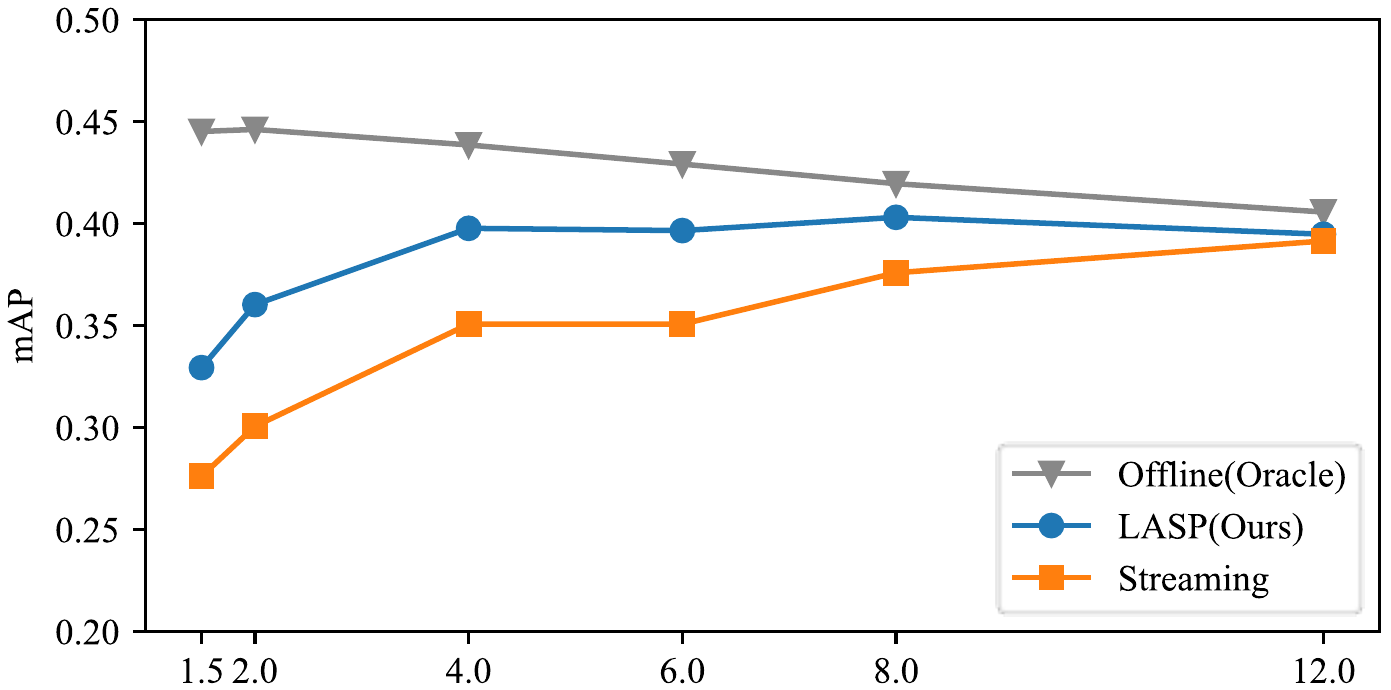}
  \vspace{-0.8cm}
\end{center}
  \caption{Online performance under different streaming frame rate}
\label{fig:streaming-rate}
\vspace{-0.4cm}
\end{figure}

\begin{table}[ht]
  \centering
  \caption{Streaming Performance on different computation platform}
  \begin{tabular}{cccccc}
      \toprule[1pt]
      GPU & Methods & FPS & mAP$\uparrow$ & NDS$\uparrow$ & mATE$\downarrow$\\
      \midrule[0.5pt]
      \multirow{6}*{RTX3090} & FCOS3D\cite{wang2021fcos3d} & 1.7 & 0.208 & 0.326 & 0.828\\
      ~                     & BEVDet\cite{huang2021bevdet} & 12.6 & 0.289 & 0.370 & 0.730\\
      ~                     & BEVFormer\cite{li2022bevformer} & 2.4 & 0.310 & 0.452 & 0.760\\
      ~                     & BEVDepth\cite{li2023bevdepth} & 8.6 & 0.323 & 0.464 & 0.654\\
      ~                     & StreamPETR\cite{wang2023exploring} & \textbf{26.1} & 0.391 & 0.493 & 0.687\\
      ~                     & LASP(Ours) & 24.3 & \textbf{0.403} & \textbf{0.506} & \textbf{0.667}\\
      \midrule[0.5pt]
      \multirow{6}*{GTX1060} & FCOS3D\cite{wang2021fcos3d} & 0.3 & 0.051 & 0.234 & 0.858\\
      ~                     & BEVDet\cite{huang2021bevdet} & 3.3 & 0.254 & 0.348 & 0.751\\
      ~                     & BEVFormer\cite{li2022bevformer} & 0.3 & 0.074 & 0.311 & 0.819\\
      ~                     & BEVDepth\cite{li2023bevdepth} & 1.4 & 0.226 & 0.404 & 0.686\\
      ~                     & StreamPETR\cite{wang2023exploring} & \textbf{6.4} & 0.368 & 0.489 & 0.693\\
      ~                     & LASP(Ours) & 5.9 & \textbf{0.398} & \textbf{0.511} & \textbf{0.655}\\
      
      \midrule[0.5pt]
      \multirow{2}*{Jetson Orin} & StreamPETR\cite{wang2023exploring} & \textbf{2.1} & 0.299 & 0.455 & 0.712\\
      ~                          & LASP(Ours) & 1.8 & \textbf{0.353} & \textbf{0.490} & \textbf{0.671}\\
      \bottomrule[1pt]
  \end{tabular}
  \label{tab:computation-platform}
  \vspace{-0.2cm}
\end{table}

\begin{figure*}[t]
  \centering
  \includegraphics[width=\textwidth]{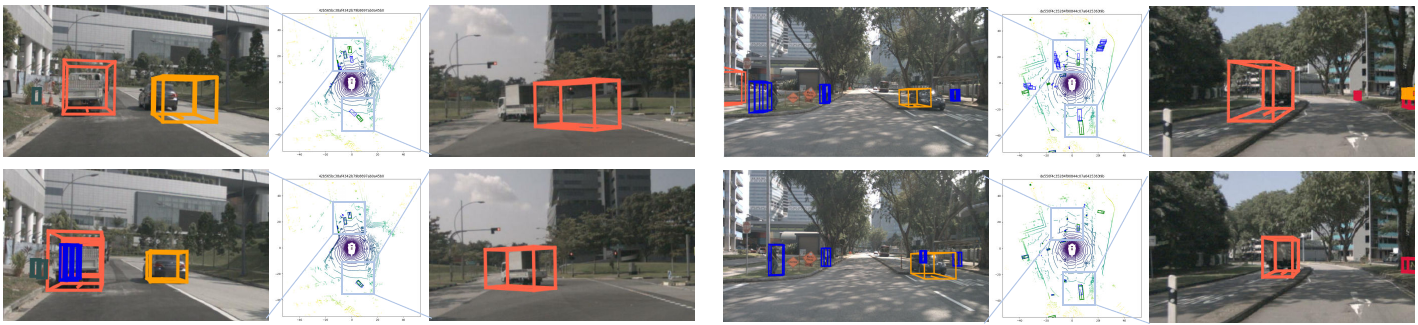}
  \vspace{-0.8cm}
  \caption{\textbf{Visualization of baseline and our results on nuScenes dataset}. We show 3D bboxes predictions in camera images and the bird’s-eye-view.}
  \label{fig:qualitative}
  \vspace{-0cm}
\end{figure*}

\begin{figure*}[t]
  \centering
  \includegraphics[width=\textwidth]{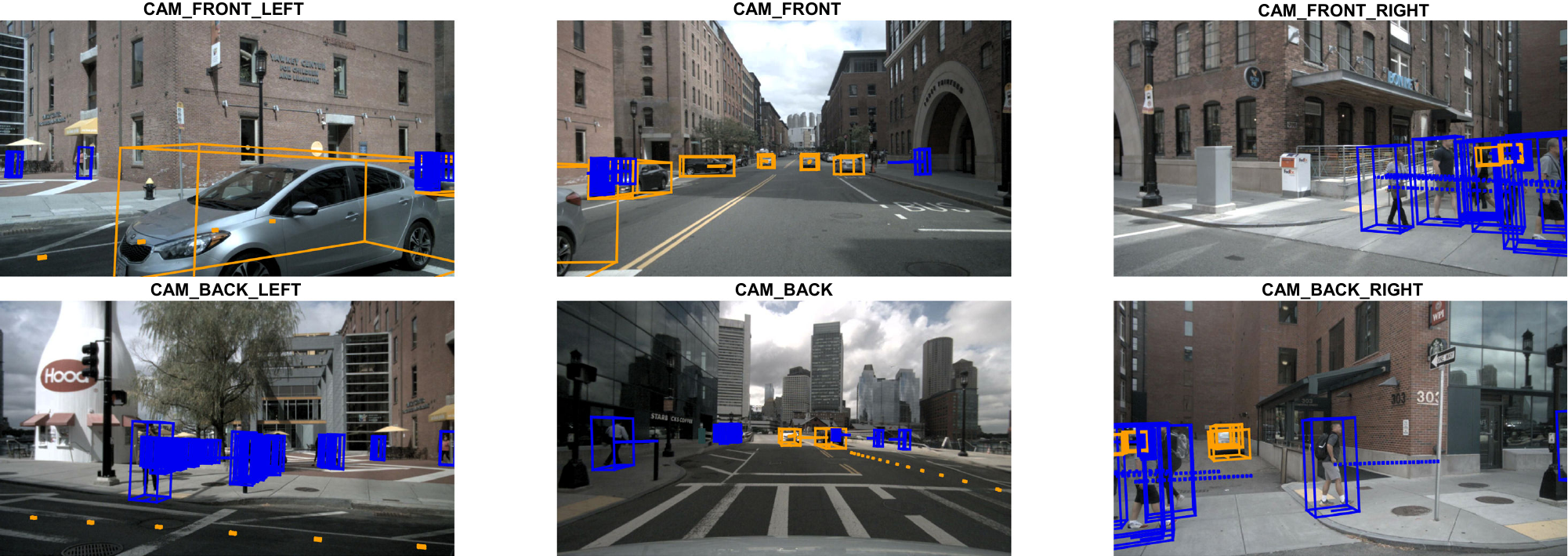}
  \vspace{-0.7cm}
  \caption{\textbf{Visualization of predictive detection results}. We show detection boxes for each object and their future trajectories.}
  \label{fig:qualitative}
  \vspace{-0.3cm}
\end{figure*}

First, we compare our framework with different simulated streaming frame rate on the established benchmark, where the model is provided with historical frame data at latency intervals $\tau$. We then evaluate the model's predictions for the next frame run period. Only a single model is used across all simulated streaming frame rates. As shown in Fig. \ref{fig:streaming-rate}, in offline condition, our model exhibits robust performance under different streaming rate inputs, capable of integrating historical data collected with irregular interval based on the inherent data timestamp. Under online streaming condition, our model demonstrates varying degrees of improvement in mAP compared to the baseline. This indicates that our approach effectively integrates historical information and exhibits strong predictive capabilities by compensating for object motion during algorithm execution, with the degree of compensation increasing as the latency grows.
Notably, for the first time, we incorporate trajectory prediction results into the evaluation of 3D perception tasks.

In order to better align with real-world scenarios, we also test our method on various computational devices in TABLE \ref{tab:computation-platform}, including the high-performance RTX 3090 and embedded system like the Jetson AGX Orin. 
Due to the streaming frame rate of the RTX 3090 exceeding the data frame rate of camera sweeps, and the minimal difference between consecutive frames beyond 12Hz, we opt to use all frames for inference. We retrained the SreamPETR\cite{wang2023exploring} model with video input at the streaming frame rate of each platform for a fair comparison. 
As the computational capacity decreases from 35.6 TFLOPS on the RTX 3090 to 5.325 TFLOPS on the Jetson AGX Orin, the algorithm's latency increases, and the time lag effect becomes more pronounced. Despite these challenges, our algorithm achieves a good balance between speed and accuracy, and demonstrates robust performance across various computational platforms. Our method achieves a 35.3\% online mAP on the Jetson AGX Orin without employing any acceleration techniques, outperforming other algorithms even on more powerful platforms.

\subsection{Ablation Study}
Next, we conduct ablation studies on the two proposed modules in Jetson AGX Orin for online evaluation.

\noindent \textbf{Latency-aware history integration: }For our latency-aware history integration module, we evaluate the model's online performance under real-world settings where random latency is sampled from Jetson AGX Orin. We compare our method against two baselines: input directly without any alignment operation, and using the Motion-aware Layer Normalization(MLN) module proposed in \cite{wang2023exploring}. Our module implicitly encodes motion attributes and explicitly models the dynamic evolution of object features over time when processing historical feature at irregular intervals, thus addressing temporal discrepancies between consecutive frames. As indicated in the TABLE \ref{tab:temporal-alignment}, our method demonstrates an advantage in handling irregularly historical data series.

\begin{table}[!t]
  \centering
  \vspace{-0.0cm}
  \caption{The ablation study on temporal alignment}
  \begin{tabular}{lccccc}
      \toprule[1pt]
      Methods & mAP$\uparrow$ & mATE$\downarrow$ & mASE$\downarrow$ & mAOE$\downarrow$ & mAAE$\downarrow$\\
      \midrule[0.5pt]
      No align & 0.327 & 0.713 & 0.272 & 0.442 & \textbf{0.212}\\
      \midrule[0.5pt]
      MLN & 0.343 & 0.704 & 0.276 & 0.462 & 0.223 \\
      \midrule[0.5pt]
      Ours & \textbf{0.353} & \textbf{0.671} & \textbf{0.270} & \textbf{0.429} & 0.217 \\
      \bottomrule[1pt]
  \end{tabular}
  \label{tab:temporal-alignment}
  \vspace{-0.4cm}
\end{table}

\noindent \textbf{Latency-aware predictive detection:} We further analyze the role of the predictive detection module in motion compensation, as shown in TABLE \ref{tab:motion-compensation}. The "zero hold" method simply copies the most recent detection results to the current query time step. Based on the uniform motion assumption, "velocity-based" compensation can be applied to estimate object movement. We also compare our method with the TensorRT-accelerated model, which achieves a frame rate (FPS) of 3.4, to emphasize the advantages of streaming perception. Furthermore, we adopt the 2D streaming "forecasting" paradigm by using future frame labels at time $t+\tau$ for supervision during training, following \cite{yang2022real}. However, this approach is limited to predicting object states at a specific time point. In contrast, our method is capable of predicting object states over a range of time steps, while simultaneously modeling interactions between objects, allowing for the prediction of non-linear movement patterns.

\begin{table}[!t]
  \centering
  \vspace{-0.0cm}
  \caption{The ablation study on motion compensation}
  \begin{tabular}{lccccc}
      \toprule[1pt]
      Methods & mAP$\uparrow$ & mATE$\downarrow$ & mASE$\downarrow$ & mAOE$\downarrow$ & mAAE$\downarrow$\\
      \midrule[0.5pt]
      Zero hold & 0.304 & 0.680 & 0.278 & 0.461 & 0.235\\
      \midrule[0.5pt]
      Forecasting & 0.314 & 0.683 & 0.272 & 0.447 & 0.236\\
      \midrule[0.5pt]
      Velocity-based & 0.331 & 0.683 & 0.272 & 0.438 & 0.228\\
      \midrule[0.5pt]
      TensorRT & 0.334 & 0.685 & 0.273 & 0.452 & 0.234\\
      \midrule[0.5pt]
      Ours & \textbf{0.353} & \textbf{0.671} & \textbf{0.270} & \textbf{0.429} & \textbf{0.217} \\
      \bottomrule[1pt]
  \end{tabular}
  \label{tab:motion-compensation}
  \vspace{-0.4cm}
\end{table}

%% file: 6.conclusion.tex
\section{conclusion}

This paper addresses the critical challenge of integrating latency awareness into 3D perception benchmark and algorithms for autonomous driving. Our proposed Latency-Aware 3D Streaming Perception (LASP) framework, built upon a query-based 3D object detectors, extends to continuous history integration and predictive detection to compensate for arbitrary algorithm delays. The framework demonstrates generalization across different latency conditions and achieving an online performance that closely aligns with 80\% of its offline evaluation on the Jetson AGX Orin without any acceleration techniques, advancing the field towards real-time, latency-resilient perception in autonomous driving.